\documentclass[twoside]{article}

\usepackage[preprint]{aistats2027}
\usepackage[round]{natbib}

\usepackage[hyphens]{url}
\usepackage{hyperref}
\hypersetup{
    colorlinks,
    citecolor=blue,
    filecolor=blue,
    linkcolor=blue,
    urlcolor=blue,
}

\usepackage{makecell}
\usepackage{subcaption}

\usepackage{graphicx}

\usepackage{amsmath}
\usepackage{amssymb}
\usepackage{amsfonts}
\usepackage{amsthm}
\usepackage{dsfont}
\usepackage{thmtools, thm-restate}

\usepackage[normalem]{ulem}

\newcommand{\Chol}[0]{\operatorname{Chol}}
\newcommand{\cov}[0]{\operatorname{cov}}
\newcommand{\diag}[0]{\operatorname{diag}}
\newcommand{\KL}[2]{D_{\mathrm{KL}}\left(#1 \middle\| #2\right)}

\newcommand{\tr}[0]{\operatorname{tr}}

\newtheorem{theorem}{Theorem}

\newtheorem{corollary}[theorem]{Corollary}

\newcommand{\hypers}{\sigma}
\newcommand{\qdiag}{\text{D}}
\newcommand{\qlrd}{\text{LRD}}
\newcommand{\qlr}{\text{LR}}
\newcommand{\qfullrank}{\text{FR}}

\begin{document}


\twocolumn[

\aistatstitle{Occam's Razor is Only as Sharp as Your ELBO}

\aistatsauthor{Ethan Harvey \And Geoff Pleiss \And Michael C. Hughes}

\aistatsaddress{Tufts University\\\texttt{ethan.harvey@tufts.edu} \And University of British Columbia\\\texttt{geoff.pleiss@stat.ubc.ca} \And Tufts University\\\texttt{michael.hughes@tufts.edu}}
]

\begin{abstract}
The marginal likelihood, also known as the \emph{evidence}, is regarded as a mathematical embodiment of Occam's razor, enabling hyperparameter selection that avoids overfitting. The \emph{evidence lower bound} (ELBO) objective from variational inference has also been used for similar purposes. In this paper, we define a signal-to-noise ratio (SNR) for Bayesian linear regression in terms of key hyperparameters controlling signal and noise variance. We then study how ELBO-based hyperparameter learning can under or overestimate SNR depending on the assumed covariance matrix structure in a Gaussian approximate posterior. We prove that assuming a diagonal or low-rank plus diagonal covariance matrix underestimates the SNR, while a low-rank covariance matrix overestimates the SNR. Underestimating the SNR leads to higher than optimal mean squared error on the training set (underfitting), while overestimating SNR leads to lower error (overfitting). Bayesian practitioners hoping to scale to large models should be cautious about how covariance assumptions needed for tractability may impact model selection.
\end{abstract}
% Keywords: Bayesian model selection, variational inference, linear algebra
% TL;DR: We prove that, in Bayesian linear regression, assuming a diagonal or low-rank plus diagonal covariance matrix underestimates the signal-to-noise ratio (SNR), while assuming a low-rank covariance matrix overestimates the SNR.
\section{INTRODUCTION}
\label{sec:introduction}

Bayesian model selection relies on the marginal likelihood, commonly referred to as the \emph{evidence}, to quantify how well a model explains the observed data while accounting for model complexity.
The evidence is regarded as a mathematical embodiment of Occam's razor, favoring simpler models that explain the data well~\citep{jeffreys1939theory,mackay1991bayesian,rasmussen2000occam,grunwald2005tutorial}.
In practice, computing the marginal likelihood exactly is intractable for many modern models, particularly deep neural networks (DNNs). This has led to the adoption of variational inference (VI)~\citep{jordan1999introduction,blei2017variational} for scalable approximate learning.% inference technique.

\begin{figure}[!t]
    \centering
    \includegraphics[width=\linewidth]{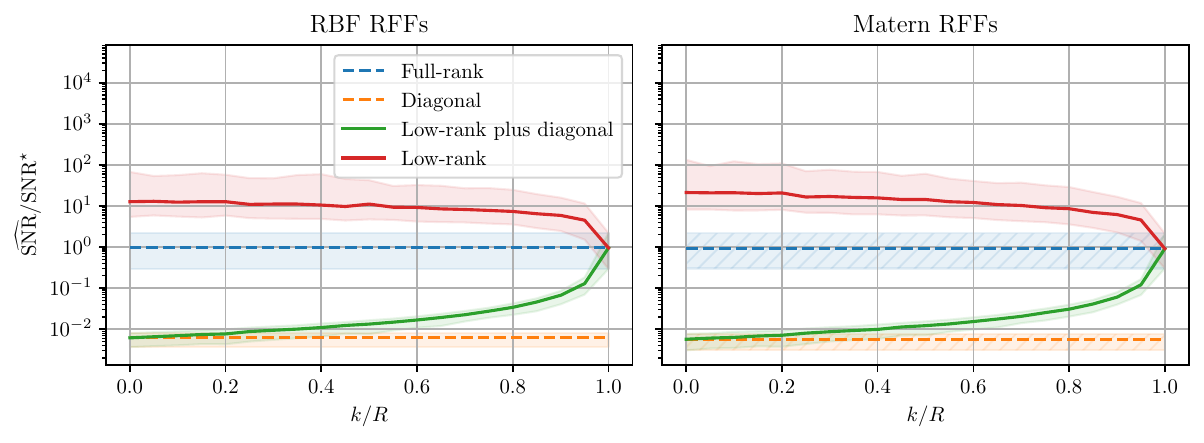}
    \caption{    
    Rank of the approximate posterior covariance $k \in (0, 1{,}024)$ divided by the number of features $R$ (x-axis) vs. estimated $\widehat{\text{SNR}}$ divided by the data-generating $\text{SNR}^\star$ (y-axis, $\text{ideal}=1$).
    Task: Bayesian linear regression on $N=20$ synthetic data points (Sec.~\ref{sec:expts}).
    We estimate the approximate posterior's mean, covariance, and hyperparameters.
    \textbf{The ELBO can under or overestimate the SNR depending on the assumed covariance matrix structure.}
    }%endcaption
    \label{fig:varying_k}
\end{figure}

\let\thefootnote\relax\footnotetext{\hspace{-20pt} Code: \href{https://github.com/ethanharvey98/Occams-razor}{\texttt{github.com/ethanharvey98/Occams-razor}}}

Given a joint model over weights $w$ and observed labels $y$, VI approximates the true posterior $p(w | y)$ with a simpler distribution $q(w)$. The parameters defining $q$ 
are optimized to maximize a tractable bound on the log evidence, the \emph{evidence lower bound} (ELBO).
The ELBO is often described as giving rise to Occam's razor~\citep{hinton1993keeping,barber1998ensemblea,barber1998ensembleb}.
% NOTE: See 5.7 Bayesian Neural Networks in Bishop (2006).
Past work has used the ELBO for model selection in mixture models~\citep{ueda2002bayesian,mcgrory2007variational}, Gaussian processes (GPs)~\citep{titsias2009variational,damianou2013deep,hensman2015scalable,lalchand2022sparse}, and DNNs~\citep{kc2021joint,wuDeterministicVariationalInference2019}. 

However, ELBO-based hyperparameter learning does not always provide an Occam's razor effect for model selection.
The tightness of this bound to the ideal log evidence depends on the chosen variational family for $q(w)$.
A user can choose the covariance structure of $q$ to balance fidelity and efficiency~\citep{challisGaussianKullbackLeiblerApproximate2013}.
As models get larger, full-rank covariances become impractical, so low-rank~\citep{seegerGaussianCovarianceScalable2010} or diagonal assumptions are needed.
How the chosen covariance structure impacts model selection, especially the ability to jointly learn an approximate posterior and hyperparameters, needs more study.
Prior work has shown that assuming a Gaussian approximate posterior with a diagonal covariance can lead to underfitting~\citep{coker2022wide,harvey2024learning,harvey2026learning}.

In this paper, we examine ELBO-based hyperparameter learning for a popular yet analytically tractable model: Bayesian linear regression. We first show that a notion of signal-to-noise ratio (SNR) involving key hyperparameters can help characterize training set fit quality (Sec.~\ref{sec:snr}). 
We then show that VI for BLR can under or overestimate the signal-to-noise ratio (SNR) depending on the assumed covariance matrix structure in a Gaussian approximate posterior.
Assuming a diagonal or low-rank plus diagonal covariance matrix underestimates the SNR (Sec.~\ref{sec:underestimating_the_snr}), while a low-rank covariance matrix overestimates the SNR (Sec.~\ref{sec:overestimating_the_snr}).
Bayesian practitioners hoping to scale to large models should be cautious about how assumptions needed for tractability may impact the potential for model selection.

\section{BACKGROUND}

We have a training set $\{(x_i, y_i)\}_{i=1}^N$ where $x_i \in \mathbb{R}^D$ denotes an input vector and $y_i \in \mathbb{R}$ denotes a scalar output or target.
Let $\phi(\cdot) : \mathbb{R}^D \rightarrow \mathbb{R}^R$ be a feature map that defines the design matrix $\Phi = [\phi(x_1)^\top; \dots; \phi(x_N)^\top] \in \mathbb{R}^{N\times R}$ and $y = [y_1; \dots; y_N] \in \mathbb{R}^N$ denote the target vector.

\textbf{Prior.}
We assume a Bayesian linear regression model with parameter vector $w \in \mathbb{R}^R$.
The prior over parameters is a zero-mean Gaussian,
\begin{align}
    p(w) = \mathcal{N}(w \mid 0_R, \Sigma_p)
\end{align}
where $0_R$ denotes the $R$-dimensional zero vector and $\Sigma_p$ is an $R \times R$ covariance matrix.
For simplicity, throughout this work we assume the prior covariance is the $R \times R$ identity matrix $\Sigma_p = I_R$.
In Sec.~\ref{sec:empirical_bayes}, we relax this assumption and learn an isotropic prior covariance via empirical Bayes.

\textbf{Likelihood.}
Given the parameter vector $w$, the observations are generated by a Gaussian likelihood,
\begin{align}
    p(y | w) = \mathcal{N}(y \mid \sigma_f \Phi w, \sigma_n^2 I_N)
\end{align}
where $\sigma_f$ denotes the signal amplitude and $\sigma_n^2$ denotes the noise variance. %\citet{rasmussen2006gaussian} define $\sigma_f^2$ as the variance of the (noise-free) signal.
We denote these two hyperparameters as $\sigma = \{\sigma_f, \sigma_n\}$.
%Later we estimate $\sig$ that needs to be estimated from data.

This likelihood implies an additive noise model, $y = f + \varepsilon$, where the \emph{signal} vector $f = \sigma_f \Phi w$ sets a mean corrupted by zero-mean isotropic \emph{noise} $\varepsilon \sim \mathcal{N}(0_N, \sigma_n^2 I_N)$.

\textbf{Posterior.}
Due to conjugacy, the exact posterior over $w$ is also Gaussian. Given hyperparmeters $\sigma$, we have
\begin{align}
    p(w | y) = \mathcal{N}(w \mid \mu^*, \Sigma^*),
    \label{eq:posterior}
\end{align}
with $\mu^* = \tfrac{\sigma_f}{\sigma_n^2} \Sigma^* \Phi^\top y$ and $\Sigma^* = (\tfrac{\sigma_f^2}{\sigma_n^2} \Phi^\top \Phi + \Sigma_p^{-1})^{-1}$.
Let $\Lambda^* = (\Sigma^*)^{-1}$ be the exact posterior precision matrix.

\textbf{Marginal Likelihood.}
Due to the compositional properties of the Gaussian, integrating out $w$ yields the marginal of $y$, often called the \emph{evidence},
\begin{align}
    \label{eq:marginal_likelihood}
    p(y) = \mathcal{N}(y \mid 0_N, \sigma_f^2 \Phi \Sigma_p \Phi^\top + \sigma_n^2 I_N).
\end{align}
\textbf{Approximate Posterior.}
VI approximates the true (intractable) posterior $p(w | y)$ with a simpler distribution $q(w)$.
We choose $q(w)$ to be Gaussian,
\begin{align}
    q(w) = \mathcal{N}(w \mid m, S),
\end{align}
and consider different possible constraints on the covariance matrix $S$.
When there are no constraints on $S$, which we call the full-rank (FR) case, the exact posterior $p(w|y)$ is realizable within this variational family.
However, it is common in practice to trade off fidelity for efficiency by restricting $S$ to some structured parametric form, such as the diagonal, low-rank plus diagonal (LRD), or low-rank (LR) we study later.

\textbf{Evidence Lower Bound.}
The ELBO objective $J$ is a function of variational parameters $m, S$ and the hyperparameters $\hypers$. We define $J$ as
\begin{align}
    \label{eq:evidence_lower_bound}
    J(m, S, \hypers) = \mathbb{E}_{q}\left[ \log p(y | w) \right] - \KL{q(w)}{p(w)}.
\end{align}
A bound relation holds: $J(m, S, \hypers) \leq \log p(y)$. 

\textbf{Variational EM.}
%Since the expected log-likelihood is the expectation of a quadratic form under a Gaussian and the KL divergence between two Gaussians has a closed-form expression, the ELBO can be computed in closed form (see App.~\ref{app:closed-form_elbo}).
We can use coordinate ascent for variational expectation-maximization (CA-VEM)~\citep[Ch.~10]{bishop2006properties} to perform iterative updates to the variational parameters $m, S$ and hyperparameters $\sigma$. Each update is guaranteed to maximize ELBO for one variable (``coordinate'') given the rest. These guarantees enable our later theoretical analysis.
The optimal updates (derived in App.~\ref{app:coordinate_ascent_updates}) are:
\begin{align}
    \label{eq:cavi_m_S}
    m(\hypers) &= \left( \tfrac{\sigma_f^2}{\sigma_n^2} \Phi^\top \Phi + \Sigma_p^{-1} \right)^{-1} \tfrac{\sigma_f}{\sigma_n^2} \Phi^\top y, \\
    S(\hypers) &= \left( \tfrac{\sigma_f^2}{\sigma_n^2} \Phi^\top \Phi + \Sigma_p^{-1} \right)^{-1}, \notag \\
    \sigma_f(m, S) &= \frac{y^\top \Phi m}{m^\top \Phi^\top \Phi m + \tr(\Phi^\top \Phi S)}, \label{eq:cavi_sigma} \\
    \sigma_n^2(m, S, \sigma_f) &= \tfrac{1}{N} \left( \left\| y - \sigma_f \Phi m \right\|_2^2 + \sigma_f^2 \tr(\Phi^\top \Phi S) \right). \notag
\end{align}
These optimal updates to $m$ and $\sigma$ hold for \emph{any} covariance assumptions, while the $S$ update above assumes the unrestricted full-rank (FR) case. 
In the FR case, we know that
$S(\sigma) {=} \Sigma^*$,  recovering the exact posterior covariance defined below Eq.~\eqref{eq:posterior}. We also know that the FR ELBO bound is tight for the current $\sigma$ immediately after updates to $m$ and $S$. Thus, in the FR case CA-VEM will have the same optimal $\sigma$ as directly maximizing the log marginal likelihood (LML).

\newcommand{\ftest}{f_*}
\newcommand{\ytest}{y_*}
\newcommand{\xtest}{x_*}
\newcommand{\Phitest}{\Phi_*}

\textbf{Predictive Variance.} Given $M$ query points $\xtest$ with features $\Phitest \in \mathbb{R}^{M \times R}$, the predictive distribution~\citep[Ch. 2]{rasmussen2006gaussian} for the corresponding noise-free latent signal $\ftest$ is
\begin{align}
    p(\ftest | y) = \mathcal{N}(\ftest \mid \sigma_f \Phitest \mu^*, \sigma_f^2 \Phitest \Sigma^* \Phitest^\top)
\end{align}
We define the noise-free predictive variance as
\begin{align}
    \mathbb{V}_{p(\ftest | y)}[\ftest] = \tr(\cov(\ftest)) = \sigma_f^2 \tr(\Phitest \Sigma^* \Phitest^\top).
\end{align}
Key steps in our later analyses involve the term $\sigma^2_f \text{tr}(\Phi^\top \Phi S)$, which we call the \emph{training set predictive variance} for an approximate posterior with covariance $S$.
We call this ``scale-free'' when we omit $\sigma^2_f$ in front.

\section{ANALYZING MODEL FIT WITH THE SIGNAL-TO-NOISE RATIO}
\label{sec:snr}

For regression model introduced above, we define its signal-to-noise ratio as
% Given a training set of size $N$, its target vector $y$ can be decomposed into a sum of a signal vector $f$ and a noise vector: $y = f + \varepsilon$, where $f = \sigma_f \Phi w$ and $\varepsilon \sim \mathcal{N}(0_N, \sigma_n^2 I_N)$.
\begin{align}
    \mathrm{SNR} \triangleq \frac{\mathbb{E}_{p(w)}[f^\top f]}{\mathbb{E}_{p(\varepsilon)}[\varepsilon^\top \varepsilon]} = \frac{\sigma_f^2 \tr(\Phi^\top \Phi \Sigma_p)}{N \sigma_n^2}.
\end{align}
This SNR definition comes from \citet[Eq. 4.141]{murphy2012SNR}, and aligns with the notion of a ``model-based'' SNR~\citep[Eq. 6]{soch2018SNR}.
\citet{welvaertDefinitionSignalToNoiseRatio2013} survey alternative SNR definitions.

Why do we care about the SNR?
The SNR gives us a single number representing how much of $y$'s variance comes from the signal versus noise.
Because the signal variance $\sigma_f^2$ and noise variance $\sigma_n^2$ are learned via the ELBO, a biased approximate posterior can push the estimated $\widehat{\mathrm{SNR}}$ away from the true $\mathrm{SNR}^*$.
As we show in the theorem below, underestimating this ratio looks like underfitting, and overestimating it looks like overfitting.
Consequently, SNR error gives us a quantitative way to diagnose over and underfitting that traces directly back to the assumed covariance structure.

\begin{restatable}[Lower SNR implies higher mean squared error on the training set.]{theorem}{thmSNRtoMSE}
    \label{theorem:SNRtoMSE}
    Let $\hat{\sigma}$ and $\sigma^*$ be two hyperparameter values, such that the SNR of the former is less than or equal to the latter:
    $\frac{\hat{\sigma}_f^2}{\hat{\sigma}_n^2} 
    \leq 
    \frac{(\sigma^*_f)^2}{(\sigma^*_n)^2} 
    $.
    Then the training set mean squared error (MSE) of the former is greater than or equal to the latter, when both use a CA-VEM update to $m$ from their respective $\sigma$ values:
    \begin{align}
        \tfrac{1}{N} \left\| y - \hat{\sigma}_{f} \Phi m(\hat{\sigma}) \right\|_2^2 \ge \tfrac{1}{N}  \left\| y - \sigma_f^* \Phi m(\sigma^*) \right\|_2^2.
    \end{align}
    This inequality is strict unless the SNRs are equal, assuming for regularity the updated means are not all zero, which just requires $y \notin \textnormal{Null}(\Phi^\top)$.
\end{restatable}

\textit{Proof Sketch.}
First, we substitute the definition of $m(\sigma)$ from Eq.~\eqref{eq:cavi_m_S} on both sides and apply the push-through identity~\citep{henderson1981deriving} to express each MSE in terms of the kernel matrix $\Phi \Phi^\top$.
Second, we apply eigendecomposition to the kernel matrix to obtain a sum over nonnegative eigenvalues.
Finally, the MSE inequality follows directly from the SNR inequality.
See proof in App.~\ref{app:proof_SNR_to_MSE}.

\section{UNDERESTIMATING THE SNR}
\label{sec:underestimating_the_snr}

Here, we show how two popular efficient parameterizations of the covariance matrix of $q(w)$ lead to underestimating the SNR, with consequent impacts on training set error that imply underfitting.

\subsection{Diagonal Covariance}
Consider restricting $q(w)$ to a diagonal (D) covariance,
\begin{align}
    q(w) = \mathcal{N}(w \mid m, \diag(s)),
\end{align}
where $\diag(s)$ is a diagonal matrix whose main diagonal equals the elements of vector $s \in \mathbb{R}_+^R$.
This is often called a ``mean-field'' approximation and is a popular choice for VI with DNNs~\citep{blundell2015weight,wuDeterministicVariationalInference2019,coker2022wide} for its affordability.

\begin{theorem}[{Optimal $s$ for diagonal VI \citep[Sec.~10.1.2]{bishop2006properties}}]
    \label{theorem:optimal_s_for_diag}
    Given hyperparameters $\sigma$ and assuming a Gaussian approximate posterior with a diagonal covariance, the ELBO is maximized by setting
    \begin{align}
        \hat{s}_r = \frac{1}{\Lambda_{rr}^*} \quad \text{for $r=1, \dots, R$}.
    \end{align}
    Here, $\Lambda^* = (\Sigma^*)^{-1}$ denotes the exact posterior precision given hyperparameters $\sigma$.
\end{theorem}
    
\textit{Proof.}
Take the partial derivative of the ELBO in Eq.~\eqref{eq:evidence_lower_bound} with respect to $s_r$, set it equal to zero, and solve to get $\hat{s}_r = \frac{1}{\Lambda_{rr}^*}.$ The second derivative confirms this is a maxima.~$\blacksquare$

\begin{restatable}[Diagonal VI overestimates the training set predictive variance]{theorem}{diagUnderestimatesPredvar}
    \label{theorem:diag_overestimates_predvar}
    Given hyperparameters $\sigma$, set $\hat{s}$ to maximize the diagonal ELBO.
    Then the resulting approximate posterior $q(w)$ overestimates the training set scale and noise-free predictive variance compared to the exact (FR) posterior given hyperparameters $\sigma$:
    \begin{align}
        \tr(\Phi^\top \Phi \diag(\hat{s})) \ge \tr(\Phi^\top \Phi \Sigma^*).
    \end{align}
\end{restatable}

\emph{Proof.} See App.~\ref{app:proof_diag_overestimates_predvar}.

\citet[Thm.~3.7]{odgers2026gaussian} proved this by defining the optimal approximate posterior eigenvalues as a weighted harmonic mean of the eigenvalues of the exact posterior.
By projecting the training data onto the eigenvectors of the exact posterior, they show that diagonal VI's expected predictive variance evaluated over the empirical training distribution exceeds that of the exact posterior.
We provide an alternative proof in App.~\ref{app:proof_diag_overestimates_predvar} using the standard definition of the optimal diagonal approximate posterior variance. We use the Schur-Horn theorem and a generalization of Jensen's inequality to show diagonal VI leads to overestimation.

\begin{corollary}[Diagonal VI underestimates the SNR]
\label{cor:diag_VI_hyper_inequality}
    Given hyperparameters $\sigma$ and mean $m$, perform CA-VEM updates to the covariance parameters, then $\sigma_n^2$, and then $\sigma_f$.
    %Then the resulting updates for $\sigma_f, \sigma_n^2$ (denoted with hats) are bounded relative to the updates using the posterior covariance (stars):
    %Next, update $\sigma_n^2$ and then $\sigma_f$.
    The resulting values (denoted with hats) for diagonal VI are bounded relative to corresponding values (stars) using FR updates:
    \begin{align}
        \hat{\sigma}^2_n \geq (\sigma^*_n)^2
        \text{~and~} 
        \hat{\sigma}_f^2 \leq (\sigma^*_f)^2,
    \end{align}
where equality holds only if $\Sigma^*$ is a diagonal matrix.
Thus, diagonal VI will underestimate the SNR after each update compared to full-rank (FR) VI.
\end{corollary}

\textit{Proof.}
We update $\sigma_n^2$ by plugging in the diagonal and full-rank covariance into the $\sigma_n^2$ update in Eq.~\eqref{eq:cavi_sigma}.
Comparing the former to the latter, we find
\begin{align}
    & \tfrac{1}{N} \left( \left\| y - \sigma_f \Phi m \right\|_2^2 + \sigma_f^2 \tr(\Phi^\top \Phi \diag(\hat{s})) \right) \\
    &\ge \tfrac{1}{N} \left( \left\| y - \sigma_f \Phi m \right\|_2^2 + \sigma_f^2 \tr(\Phi^\top \Phi \Sigma^*) \right), \notag
\end{align}
which holds by the inequality in Thm.~\ref{theorem:diag_overestimates_predvar}, as long as $m$ and $\sigma_f$ are the same on both sides.

Next, we consider the CA-VEM update to $\sigma_f$ in Eq.~\eqref{eq:cavi_sigma}. Comparing the diagonal to the FR updated values of $\sigma^2_f$ for the intended inequality:
\begin{align}
    &\left( \frac{y^\top \Phi m}{m^\top \Phi^\top \Phi m + \tr(\Phi^\top \Phi \diag(\hat{s}))} \right)^2 \\
    &\le \left( \frac{y^\top \Phi m}{m^\top \Phi^\top \Phi m + \tr(\Phi^\top \Phi \Sigma^*)} \right)^2. \notag
\end{align}
which again holds due to the inequality in Thm.~\ref{theorem:diag_overestimates_predvar}, when $m$ is the same on both sides.

Thus, updates to $\sigma_n$ and then $\sigma_f$ obey~$\widehat{\text{SNR}}_{\qdiag} {\le} \widehat{\text{SNR}}_{\qfullrank}$, because SNR is a positive multiple of  $\sigma_f^2 / \sigma_n^2.~\blacksquare$

Later in Sec.~\ref{sec:residual_error_when_SNR_underest}, we show that this underestimation of SNR has serious consequences for prediction quality: it leads to higher mean squared error on the training set than a full-rank posterior.

\subsection{Low-Rank Plus Diagonal Covariance}

Consider restricting $q(w)$ to a low-rank plus diagonal ($\qlrd$) covariance matrix,
\begin{align}
    q(w) = \mathcal{N}(w \mid m, V V^\top + D).
\end{align}
Without loss of generality, 
given a desired rank $k$, we can parameterize the low-rank plus diagonal covariance matrix up to a permutation as $V = \begin{bmatrix} L \\ C L \end{bmatrix}$, where $L \in \mathbb{R}^{k \times k}$ is lower triangular and $C \in \mathbb{R}^{R - k \times k}$, and $D = \diag\left( \begin{bmatrix} 0_k \\ d \end{bmatrix} \right)$, where $d \in \mathbb{R}^{R - k}$.
LRD covariance assumptions have been previously attempted for DNNs~\citep{tomczakEfficientLowRank2020,mishkinSlangFastStructured2018} to be more affordable than FR yet still more flexible than diagonal covariances. Increasing $k$ improves fidelity at the cost of scalability; $k=R$ recovers the FR case.

\begin{restatable}[Optimal $L$, $C$, and $d$ for low-rank plus diagonal VI]{theorem}{optimalParamForLRD}
\label{theorem:optimal_param_for_LRD}
    Given hyperparameters $\sigma$ and assuming a Gaussian approximate posterior with a low-rank plus diagonal covariance, the ELBO is maximized by setting
    \begin{align}
        \hat{L} &= \Chol(\Sigma_{11}^*), \quad 
        \hat{C} = \Sigma_{21}^* (\Sigma_{11}^*)^{-1}, \\
        \hat{d}_r &= \frac{1}{(\Lambda_{22}^*)_{rr}} \quad \text{for $r=1, \dots, R-k$}. \notag
    \end{align}
    Here, $\Sigma^* = \begin{bmatrix} \Sigma_{11}^* & \Sigma_{12}^* \\ \Sigma_{21}^* & \Sigma_{22}^* \end{bmatrix}$ and $\Lambda^* = \begin{bmatrix} \Lambda_{11}^* & \Lambda_{12}^* \\ \Lambda_{21}^* & \Lambda_{22}^* \end{bmatrix}$ denote the block partitions of the exact covariance and precision matrices given hyperparameters $\sigma$.
\end{restatable}

\textit{Proof Sketch.} 
We decompose $q(w_1, w_2) = q(w_1) q(w_2 | w_1)$ and solve the KL minimization problem equivalent to maximizing the low-rank plus diagonal ELBO for each factor.
Proof in App.~\ref{app:proof_optimal_param_for_LRD}. $\blacksquare$

\begin{restatable}[Low-rank plus diagonal VI overestimates the training set predictive variance]{theorem}{thmLRDOverestimatesPredVar}
    \label{theorem:LRD_overestimates_predvar}
    Given hyperparameters $\sigma$, set $\hat{L}$, $\hat{C}$, and $\hat{d}$ to maximize the low-rank plus diagonal ELBO. 
    Then the resulting $q(w)$ overestimates the training set scale and noise-free predictive variance compared to the exact (FR) posterior at that same $\sigma$:
        \begin{align}
        \tr(\Phi^\top \Phi (\widehat{VV^\top + D})) \ge \tr(\Phi^\top \Phi \Sigma^*).
    \end{align}
\end{restatable}

\textit{Proof Sketch.} Let $\Phi = \begin{bmatrix} \Phi_1 & \Phi_2 \end{bmatrix}$ be the block partition of the design matrix corresponding to the block structure of the covariance matrix.
We then express the predictive variance for both the low-rank plus diagonal covariance and the FR posterior in terms of the four block parts of $\Sigma^*$. The first three terms match exactly and cancel, leaving the inequality
\begin{align}
    &\tr(\Phi_2^\top \Phi_2 \Sigma_{21}^* (\Sigma_{11}^*)^{-1} \Sigma_{12}^*) + \tr(\Phi_2^\top \Phi_2 \diag(\hat{d})) \\
    &\ge \tr(\Phi_2^\top \Phi_2 \Sigma_{22}^*). \notag
\end{align}
This holds since the diagonal component $\diag(\hat{d})$ is an approximation of the Schur complement $\Sigma_{2|1}^*$ and, by Thm.~\ref{theorem:diag_overestimates_predvar}, diagonal VI overestimates the predictive variance compared to the full-rank Schur complement.
More detailed proof in App.~\ref{app:proof_LRD_overestimates_predvar}.

\begin{corollary}[Low-rank plus diagonal VI underestimates the SNR]
\label{cor:LRD_hyperparam_inequality}
    Given hyperparameters $\sigma$ and mean $m$, perform CA-VEM updates to the covariance parameters, then $\sigma^2_n$, and then $\sigma_f$.
    The resulting values for the LRD case  (denoted with hats) are bounded relative to corresponding values (stars) using FR updates:
    \begin{align}
        \hat{\sigma}^2_n \geq (\sigma^*_n)^2,
        \text{~and~} 
        \hat{\sigma}_f^2 \leq (\sigma^*_f)^2,
    \end{align}
    where equality holds only if $\Sigma^*$ matches the covariance structure of the rank $k$ LRD family.
Thus, LRD VI will underestimate the SNR after each update compared to FR VI.
\end{corollary}

\textit{Proof sketch.} Follow the same logic as the proof of Cor.~\ref{cor:diag_VI_hyper_inequality}, but use Thm.~\ref{theorem:LRD_overestimates_predvar} to show each inequality.

\subsection{Consequent Higher Mean Squared Error}
\label{sec:residual_error_when_SNR_underest}

\begin{restatable}[Diagonal and low-rank plus diagonal VI both overestimate train set mean squared error]{corollary}{thmVIOverestimatesMSE}
    \label{cor:diag_and_LRD_overestimate_MSE}
    Given a hyperparameter value $\sigma'$, perform one full CA-VEM sweep to update $S$, then $m$, then $\sigma_n$, then $\sigma_f$, then $m$ again, using either the $\qdiag$ or $\qlrd$ approximate posterior covariance assumption.
    The resulting mean squared error (MSE) on the train set is greater than or equal to the MSE of the FR case:
    \begin{align}
        \tfrac{1}{N} 
        \left\| y - \hat{\sigma}_{f} \Phi m(\hat{\sigma}) \right\|_2^2 
        \ge 
        \tfrac{1}{N}
        \left\| y - \sigma_f^* \Phi m(\sigma^*) \right\|_2^2.
    \end{align}
    This inequality is strict unless $\Sigma^*$ matches the diagonal or LRD assumptions of $q$.
    Here, the notation $\sigma^*$ indicates the hyperparameter value after one CA-VEM sweep from $\sigma'$ using the FR posterior; $\hat{\sigma}$ denotes the same quantity after using Diagonal or LRD VI.
\end{restatable}

\textit{Proof Sketch.}
This follows from Thm.~\ref{theorem:SNRtoMSE}, with the key SNR underestimation arising from
Cor.~\ref{cor:diag_VI_hyper_inequality} for the diagonal case, or Cor.~\ref{cor:LRD_hyperparam_inequality} for the LRD case.

\textbf{Implications.}
Common folk wisdom suggests that VI with a Gaussian approximate posterior family can make reasonably accurate predictions in terms of matching the mean. Instead, its chief limitation is often understood to be underestimating uncertainty when covariance assumptions do not match the data-generating process~\citep[Fig.~10.2]{bishop2006properties},~\citep{turner2011two,giordanoLinearResponseMethods2015}.

Cor.~\ref{cor:diag_and_LRD_overestimate_MSE} shows that for simple Bayesian linear regression, if VI is used to update both $q(w)$ \emph{and} hyperparameters, then the predictions themselves (not just uncertainty) will be inaccurate, and particularly will underfit the train set. Even starting $\sigma'$ at the Bayes optimal value, the MSE will get worse after one CA-VEM sweep with either D or LRD assumptions for $q(w)$.
Prior studies where the VI mean itself does not fit data well~\citep{coker2022wide} often involve more complex models like DNNs and do not cover hyperparameter learning.

\section{OVERESTIMATING THE SNR}
\label{sec:overestimating_the_snr}

Now, we show how another possible covariance matrix parameterization can lead to overestimation of SNR and thus lower than optimal training set error. 

\subsection{Low-Rank Covariance}

Consider restricting $q(w)$ to a low-rank (LR) covariance matrix with chosen rank $k$,
\begin{align}
    q(w) = \mathcal{N}(w \mid m, V V^\top + D_{\epsilon}).
\end{align}
Without loss of generality, we can parameterize the low-rank covariance matrix up to a permutation as $V = \begin{bmatrix} L \\ C L \end{bmatrix}$, where $L \in \mathbb{R}^{k \times k}$ is lower triangular and $C \in \mathbb{R}^{R - k \times k}$.
We add a fixed small diagonal component $D_{\epsilon} = \diag([0_k; \epsilon 1_{R-k}])$ to ensure the covariance matrix is positive definite.
% Note: We are using \varepsilon for noise in the SNR and \epsilon for jitter in the low-rank covariance. \citet{rasmussen2006gaussian} use the same notation.
% \todo{TODO: Cite \citet{wenger2026variational}.}
Admittedly, while it is affordable this LR case has not yet been used in practice to our knowledge. Yet we study it here due to its surprising differences with the previous LRD case.

\begin{theorem}[Optimal $L, C$ for low-rank VI]
\label{theorem:optimal_param_for_LR}
    Given hyperparameters $\sigma$ and assuming a Gaussian approximate posterior with low-rank covariance, the ELBO is maximized by setting $L, C$ the same as Thm.~\ref{theorem:optimal_param_for_LRD}, so $\hat{L} = \Chol(\Sigma_{11}^*)$ and $\hat{C} = \Sigma_{21}^*(\Sigma_{11}^*)^{-1}$, where $\Sigma^*$ is the exact posterior covariance at the given $\sigma$ value.
\end{theorem}
\emph{Proof Sketch.} Follow the same steps as the proof of Thm.~\ref{theorem:optimal_param_for_LRD}, using a similar decomposition into $q(w_1)q(w_2|w_1)$.

\begin{theorem}[Low-rank VI underestimates the training set predictive variance]
    \label{theorem:LR_underestimates_predvar}
    Given hyperparameters $\sigma$, set $\hat{L}$ and $\hat{C}$ to maximize the low-rank ELBO.
    Then, for sufficiently small $\epsilon$, the resulting $q(w)$ \underline{underestimates} the training set scale and noise-free  predictive variance compared to the exact (FR) posterior at the same $\sigma$:
    \begin{align}
        \tr(\Phi^\top \Phi (\hat{V}\hat{V}^\top + D_\epsilon)) \le \tr(\Phi^\top \Phi \Sigma^*).
    \end{align}
\end{theorem}

\textit{Proof.}
Following the same expansion of these expressions from Thm.~\ref{theorem:LRD_overestimates_predvar} using the 4 blocks of $\Sigma^*$, the first three terms match exactly and cancel, leaving a term involving the small diagonal component $D_{\epsilon}$ for the LR case, versus the same term with the Schur complement $\Sigma^*_{2|1}$ in the FR case.
If $\epsilon \le \frac{\tr(\Phi_2^\top \Phi_2 \Sigma_{2|1}^*)}{\tr(\Phi_2^\top \Phi_2)}$, the desired inequality holds.$~\blacksquare$

\begin{corollary}[Low-rank VI overestimates the SNR]
    \label{cor:LR_overestimates_SNR}
    Given values for hyperparameters $\sigma$ and mean $m$, perform CA-VEM updates to covariance parameters, then $\sigma_n^2$ and then $\sigma_f$. The resulting values for LR VI (denoted with hats) are bounded relative to corresponding values (stars) using FR updates:
    \begin{align}
        \hat{\sigma}_n^2 \leq (\sigma_n^*)^2, \text{~and~}
        \hat{\sigma}^2_f \geq (\sigma_f^*)^2,
    \end{align}
    where equality holds only if $\Sigma^*$ matches the covariance structure of the rank $k$ LR family.
    Thus, LR VI will overestimate the SNR after each update compared to FR VI.
\end{corollary}

\textit{Proof sketch.} Follow the same logic as the proof of Cor.~\ref{cor:diag_VI_hyper_inequality}, but flip the direction of each desired inequality and use Thm.~\ref{theorem:LR_underestimates_predvar} to justify them.

\subsection{Consequent Lower Mean Squared Error}

\begin{restatable}[Low-rank VI underestimates train set mean squared error]{corollary}{thmLRVIUnderestimatesMSE}
    \label{cor:LRVI_underestimates_MSE}
    Given a hyperparameter value $\sigma'$, perform one full CA-VEM sweep to update $S$, then $m$, then $\sigma_n$, then $\sigma_f$, then $m$ again, using the $\qlr$ approximate posterior covariance assumption.
    The resulting mean squared error (MSE) on the train set is less than or equal to the MSE using the FR updates:
    \begin{align}
        \tfrac{1}{N} \left\| y - \hat{\sigma}_{f} \Phi m(\hat{\sigma}) \right\|_2^2 
        \leq 
        \tfrac{1}{N} \left\| y - \sigma_f^* \Phi m(\sigma^*) \right\|_2^2.
    \end{align}
    This inequality is strict unless $\Sigma^*$ matches the LR assumptions of $q$.
    Here, the notation $\sigma^*$ indicates the hyperparameter value after one CA-VEM sweep from $\sigma'$ using the FR posterior; $\hat{\sigma}$ denotes the same under LR VI.
\end{restatable}

\textit{Proof Sketch.}
This follows from Thm.~\ref{theorem:SNRtoMSE}, with the key SNR overestimation inequality from
Cor.~\ref{cor:LR_overestimates_SNR}.

\section{EXPERIMENTS AND RESULTS}
\label{sec:expts}

\textbf{Synthetic Data.}
To empirically validate our theoretical results, we use a synthetic data-generating process (DGP) with known SNR.
We sample inputs from a univariate ($D{=}1$) standard normal distribution, $x \sim \mathcal{N}(0, 1)$, and outputs from a GP, $y \sim \mathcal{N}(0_N, K + \sigma_n^2 I_N)$, where $K$ denotes the $N \times N$ kernel matrix induced by either a radial basis function (RBF) or Mat\'ern kernel with smoothness parameter $\nu=5/2$.
For the true DGP, we set the lengthscale $\ell {=} 1$, the signal variance $\sigma_f^2 {=} 1$, and the noise variance $\sigma_n^2 {=} 0.1^2$.
Since the data-generating signal variance $\sigma_f^2$ and noise variance $\sigma_n^2$ are known by construction, we can compare each approximate posterior's estimated $\widehat{\mathrm{SNR}}$ with the true DGP's $\mathrm{SNR}^\star$.

While the data-generating process uses the kernel function, Bayesian linear regression requires a finite-dimensional feature map $\phi(\cdot) : \mathbb{R}^D \rightarrow \mathbb{R}^R$.
We approximate the kernel function using random Fourier features (RFFs)~\citep{rahimi2007random},
\begin{align}
    \phi(x_i) = \sqrt{\tfrac{2}{R}} \cos \left( \tfrac{1}{\ell} A^\top x_i + b \right),
\end{align}
where $A \in \mathbb{R}^{D \times R}$ and $b \in \mathbb{R}^R$ are fixed weights and $\ell$ is the lengthscale.
$A$ is sampled from the spectral density of the kernel function (Gaussian for the RBF kernel and Student-$t$ for the Mat\'ern~\citep[Tab.~2]{ahirFeatureMapsLaplacian2025}) and $b$ is sampled uniformly from $[0, 2\pi)$.

As $R$ gets larger, the approximation $K \approx \Phi \Phi^\top$ gets more accurate. Thus, at large enough $R$ ELBO-based learning of a full-rank approximate posterior and hyperparameters $\sigma$ should be equivalent to directly picking $\sigma$ to optimize LML. 
Yet while the LML automatically balances data fit and model complexity, we'll see that non-full-rank ELBO inherits this Occam's razor effect only to the extent that the assumed covariance matrix structure can match the exact posterior.

\textbf{Experiment Setup.} 
We draw many N=20 datasets from this DGP. For each covariance type (D, LRD, LR, FR), we run 1000 CA-VEM iterations from a common initialization. Reproducible details in App.~\ref{app:expt_setup}.
%\section{RESULTS}

\textbf{Results and Analysis.}
Fig.~\ref{fig:varying_k} shows that both
the diagonal case and the low-rank plus diagonal case with $k < R$ underestimate the SNR.
These approximations lack the capacity to represent all correlations in the exact posterior.
In contrast, the low-rank only covariance with $k < R$, which lacks the capacity to represent uncertainty in the exact posterior outside of its dominant subspace, overestimates the SNR. This empirical under and overestimation validates our main theory in Sec.~\ref{sec:underestimating_the_snr} and~\ref{sec:overestimating_the_snr}.
Reassuringly, as the rank $k$ approaches $R$ for both LR and LRD VI, the estimated SNR becomes less biased, approaching the true SNR$^*$ at $k = R$.

\begin{figure*}[htpb!]
    \centering
    \includegraphics[width=\linewidth]{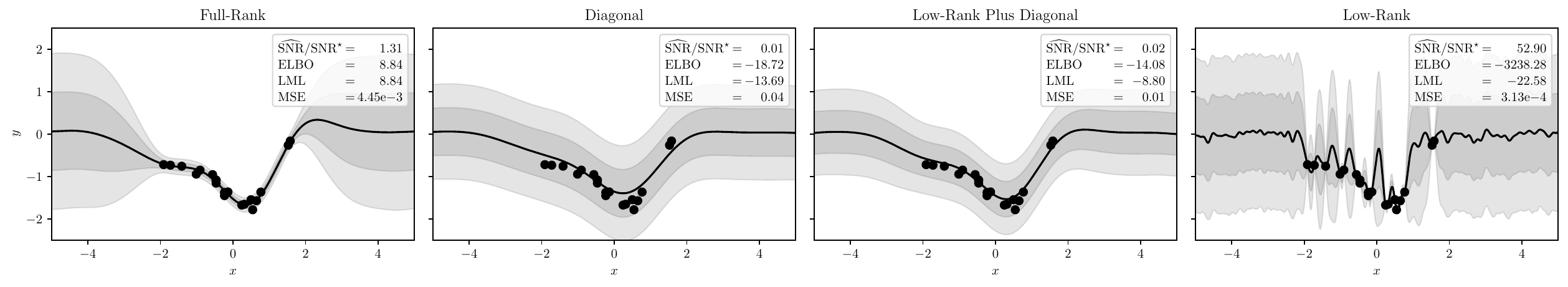}
    \caption{
    Predictive posterior $p(\ytest|y)$ for full-rank, diagonal, low-rank plus diagonal, and low-rank covariances (left to right), after estimating variational parameters and hyperparameters.
    We fix $N{=}20$, $R{=}1{,}024$, and $k{=}512$.
    }%endcaption
    \label{fig:under_overestimating}
\end{figure*}

Fig.~\ref{fig:under_overestimating} shows the same under and overfitting asymmetry in the predictive posterior $p(\ytest |y) \approx \int_w p(\ytest | w) q(w) dw$.
Both diagonal and LRD cases show rather high uncertainty even near training data.
Moreover, both cases have notably higher train set MSE than the FR case, validating Cor.~\ref{cor:diag_and_LRD_overestimate_MSE}.
In contrast, the LR only case (far right panel) shows visual overfitting to the training set, with lower train set MSE than the FR solution as predicted by Cor.~\ref{cor:LRVI_underestimates_MSE}.
As expected, the FR case (far left) shows a good fit. Its estimated SNR slightly exceeds the DGP's SNR$^*$, but this variation seems reasonable given the small training set. % of $N{=}20$.

% Surprisingly, between the under and overestimated SNRs, Bayesian model selection via the evidence (aka LML) itself sometimes prefers the overestimated SNR, while the ELBO does not.
% This could explain the empirical finding that deep kernel learning (DKL) overfits~\citep{wilson2016deep,ober2021promises,wilson2025response}, since DKL also directly optimizes hyperparameters by maximizing the LML.
% DKL learns a DNN feature map and hyperparameters by maximizing the LML.

\begin{figure*}[htpb!]
    \centering
    \includegraphics[width=\linewidth]{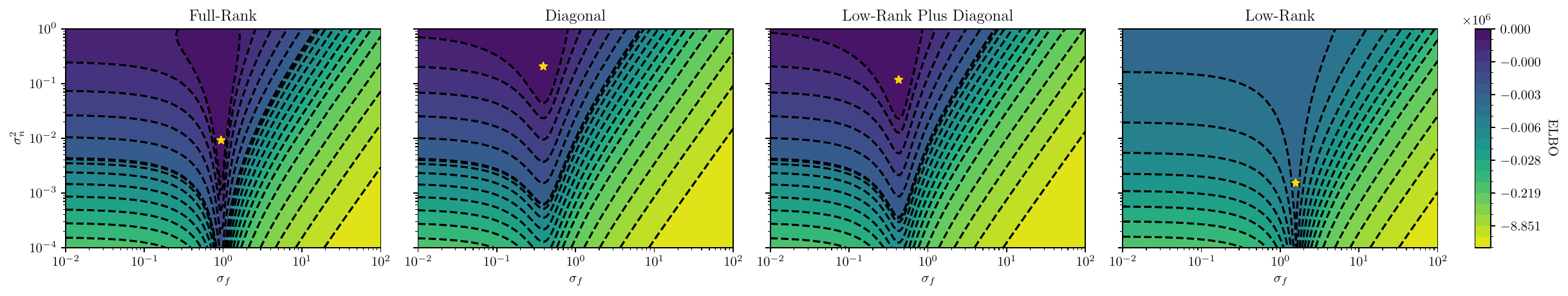}
    \caption{Contour plot showing the ELBO as a function of the signal amplitude $\sigma_f$ and noise variance $\sigma_n^2$, for the same data in Fig.~\ref{fig:under_overestimating}.
    }%endcaption
    \label{fig:varying_signal_and_noise}
\end{figure*}

Fig.~\ref{fig:varying_signal_and_noise} shows contour plots of the ELBO as a function of the signal amplitude $\sigma_f$ and noise variance $\sigma_n^2$.
The star on each panel marks the optimal $\sigma$ for each covariance structure. All ELBOs in a panel use $m,S$ values fixed to their CA-VEM updates at the starred $\sigma$.
The diagonal and low-rank plus diagonal covariances converge to a 
% MCH: to consider, should we say amplitude or variance? 
% for now, left text as amplitude since it still implies the claim about variance
%smaller signal variance $\sigma^2_f$ 
smaller signal amplitude $\sigma_f$ 
and larger noise variance $\sigma_n^2$ than the full-rank covariance, validating Cor.~\ref{cor:diag_VI_hyper_inequality} and~\ref{cor:LRD_hyperparam_inequality}.
The low-rank covariance converges to a larger signal amplitude $\sigma_f$ and smaller noise variance $\sigma_n^2$ than the full-rank covariance, validating Cor.~\ref{cor:LR_overestimates_SNR}.

\section{DISCUSSION AND RELATED WORK}
\label{sec:discussion}

\textbf{Is the prior or likelihood misspecified?}
Overfitting in Bayesian models can occur when the likelihood is narrow and the prior is broad \citep{takada2018more}.
However, the observed overfitting with LR $q(w)$ is not a result of a misspecified prior or likelihood.
% The marginal likelihood in Eq.~\eqref{eq:marginal_likelihood} is the \emph{exact same} marginal likelihood as the Gaussian process regression model with a RBF kernel, where we have latent function values $f \sim \mathcal{N}(0_N, K)$ and then observed targets $y | f \sim \mathcal{N}(f, \sigma_n^2 I_N)$, as long as the approximation $K \approx \Phi \Phi^\top$ is accurate.
Instead, our theory about under and overfitting arises from the inability of the low-rank covariance to match the exact posterior.

\textbf{Does variational inference recover the exact mean?}
\citet{margossian2024variational} proved that variational inference exactly recovers the mean of $p(w | y)$ when $p(w | y)$ exhibits an even symmetry.
This theoretical guarantee holds for the mean even when $q(w)$ is factorized and $p(w | y)$ is not.
However, this exact recovery relies on the assumption that the target posterior is fixed.
Learning hyperparameters $\sigma$ via the ELBO (type-II maximum likelihood) breaks this assumption, which is why our results differ.
%When jointly learning both $q(w)$ and $\sigma$, \citet{margossian2024variational}'s guarantees do not hold.

\textbf{What happens as dataset size increases?}
Our theory holds for a training set of fixed size $N$. Well-known asymptotic consistency results for ML estimation and VI~\citep{wangFrequentistConsistencyVariational2019} suggest that as $N$ gets larger, both estimated hyperparameters $\sigma$ and posterior $q$ should contract around data-generating point values, even when diagonal covariance assumptions are misspecified.
We thus expect the under/overfitting from our theory at finite $N$ to be alleviated with more training data; see Fig.~\ref{fig:varying_N} for empirical confirmation.

\begin{figure}[t!]
    \centering
    \includegraphics[width=\linewidth]{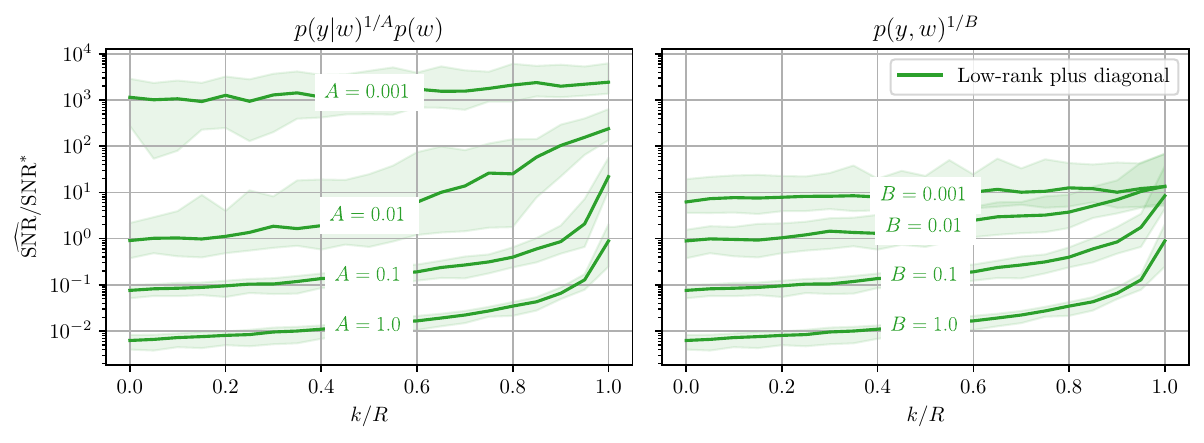}
    \caption{Estimated SNR divided by the true SNR (y-axis; ideal = 1) vs. relative approximate posterior covariance rank for a tempered likelihood (left) and joint (right). We vary $k \in (0, 1{,}024)$ and fix $N=20$ and $R=1{,}024$. \textbf{The optimal temperature depends on the flexibility of your assumed approximate posterior to match the true posterior.}
    }%endcaption
    \label{fig:varying_k_and_temp}
\end{figure}

\textbf{Can either a tempered likelihood or joint correct the underestimated SNR?}
A common approach to address miscalibration is to use a tempered likelihood $p(y | w)^{1/A} p(w)$~\citep{zhang2018noisy,osawa2019practical} or joint $p(y, w)^{1/B}$~\citep{wenzel2020good,kapoor2022uncertainty}, for positive scalars $A$ and $B$.
Since both the likelihood $p(y | w)$ and prior $p(w)$ are Gaussian in Bayesian linear regression, tempering either density by $1/T$ produces the \emph{exact same} posterior as scaling its covariance matrix by $T$.
We use this property to define the effective SNR for both a tempered likelihood and joint:
\begin{align}
    \mathrm{SNR}_A &\triangleq \frac{\sigma_f^2 \tr(\Phi^\top \Phi \Sigma_p)}{N (A \sigma_n^2)} \\
    \mathrm{SNR}_B &\triangleq \frac{\sigma_f^2 \tr(\Phi^\top \Phi (B \Sigma_p))}{N (B \sigma_n^2)} = \frac{\sigma_f^2 \tr(\Phi^\top \Phi \Sigma_p)}{N \sigma_n^2}. \notag
\end{align}
Fig.~\ref{fig:varying_k_and_temp} shows that both a tempered likelihood and joint can correct the underestimated SNR when applying LRD VI to the same synthetic task as Fig.~\ref{fig:varying_k}.
The tempered likelihood rescales the target SNR, while the tempered joint rescales the approximate posterior covariance, which in turn corrects the estimated $\widehat{\mathrm{SNR}}$.
If $A$ or $B$ is properly tuned, we can recover good $\sigma$ and thus good fits at modest rank $k$.

\textbf{How could these results extend beyond linear regression?}
We conjecture that the overestimation of the SNR for LR VI and its consequent overfitting is not unique to Bayesian linear regression.
The variational sparse GP (SVGP)~\citep{titsias2009variational} learns inducing points and hyperparameters by maximizing the ELBO.
The inducing points restrict the function-space posterior to a low-rank structure in essentially the same way as our low-rank covariance, and SVGP has been empirically shown to overfit~\citep{titsias2009variational}.

\section{EMPIRICAL BAYES RESULTS}
\label{sec:empirical_bayes}

We now set the prior covariance more flexibly as $\Sigma_p = \tau I_R$, then learn prior variance $\tau$ via type-II ML estimation along with $\sigma$ rather than fixing $\tau = 1$. This is called \emph{empirical Bayes}, and is known to prevent overfitting when maximizing the LML~\citep{mackay1992bayesian}. 
% automatic model selection

\textbf{Does empirical Bayes prevent diagonal VI from underestimating the SNR?}
No, empirical Bayes promotes the underestimated SNR rather than preventing it.
Thm.~\ref{theorem:diag_underestimates_postvar} proves that assuming a diagonal covariance underestimates the \emph{total posterior variance} (see also \citet[Lem.~A.3]{odgers2026gaussian}).
Since the total posterior variance appears in the CA-VEM update for $\tau$, the diagonal update underestimates full-rank update (see Cor.~\ref{corollary:empirical_bayes}).
This is consistent with the empirical finding in \citet[Fig.~1(a)]{harvey2026learning}: VI with isotropic covariance  underestimates the prior variance.

% \todo{TODO: Learning a prior variance causes overpruning in mean-field VI \citep{trippe2018overpruning} and biases parameters away from likelihood peaks \citep{turner2011two}}

\begin{figure}[t!]
    \centering
    \includegraphics[width=\linewidth]{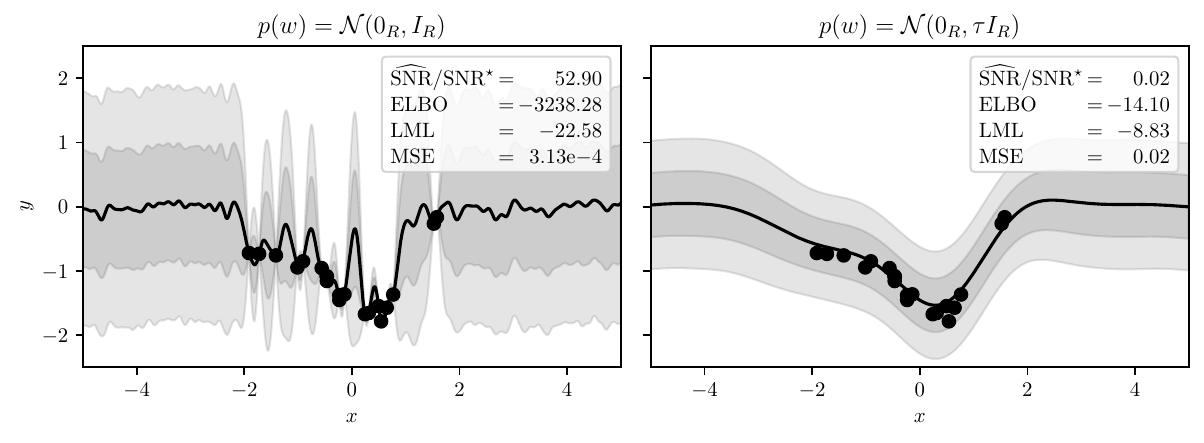}
    \caption{Predictive posterior for low-rank covariance with $p(w) = \mathcal{N}(0_R, I_R)$ (left) and $p(w) = \mathcal{N}(0_R, \tau I_R)$ (right). We fix $N=20$, $R=1{,024}$, and $k=512$.
    \textbf{Empirical Bayes prevents the low-rank covariance from overestimating the SNR.}
    }%endcaption
    \label{fig:empirical_bayes}
\end{figure}

\textbf{Does empirical Bayes prevent LR VI from overestimating the SNR?}
Fig.~\ref{fig:empirical_bayes} shows that empirical Bayes prevents the low-rank covariance from overestimating the SNR. The fix arises by driving the prior variance to the jitter $\tau \rightarrow \epsilon$. % (phew, what a relief).

\section{CONCLUSION}

We showed that in variational inference the ELBO does not always give rise to Occam's razor.
This behavior is determined by the flexibility of your assumed approximate posterior to match the exact posterior.

We hope this paper highlights the need for better marginal likelihood-based objectives that are not biased by the assumptions we need to make in order to scale to modern models.
Promising directions include modified ELBO objectives based off your assumed approximate posterior~\citep{harvey2024learning,harvey2026learning}, tighter bounds such as the importance-weighted ELBO (IWELBO)~\citep{burda2016importance} or a collapsed ELBO \citep{tomczakCollapsedVariationalBounds2021}, and more flexible divergence measures beyond KL that can balance mode-seeking and mass-covering behavior such as R\'enyi's $\alpha$-divergence~\citep{li2016renyi}.
% Kam{\'e}lia Daudel, who was a keynote speaker at the First Workshop on Advances in Post-Bayesian Methods, wrote a JMLR paper about alpha-divergence variational inference and importance weighted auto-encoders.
% Tim G. J. Rudner mentioned \citet{rainforth2018tighter} in my dissertation defense, which seems relevant for a discussion about IWELBO.

% Some quotes we could use:
% Bishop says ``there is no 'over-fitting' associated with highly flexible distributions'' for the approximate posterior $q(w)$. from Bishop Ch. 10 page 464

\textbf{Limitations.} (1) Our theory holds for just one simple model: Bayesian linear regression. Extending our theory to ELBO-based estimation of $q$ and hyperparameters for more complex regression models remains future work, especially when exact CAVI updates are no longer tractable.
(2) We focus on ELBO objectives; alternative objectives for estimating hyperparameters, such as heldout error or the conditional LML, could yield different $\sigma$ and different MSE behavior.
(3) We assume the prior covariance is identity. We conjecture that our proofs should translate to the general case, but this would require more complex reasoning and exposition.
(4) We only study a few popular covariance factorizations; we do not cover others like banded or sparse structures~\citep[Fig.~2]{challisGaussianKullbackLeiblerApproximate2013}.

\clearpage 
\subsection*{AI use statement}

In this work, we used generative AI tools for formulating mathematical claims, providing critical ingredients for proving mathematical claims, and assisting in the writing of proofs.
We have not used generative AI tools for generating synthetic data sets, helping develop theoretical models or conceptual frameworks, proposing or refining hypotheses, designing or providing feedback on research methodology or experiments, implementing methods, assisting with translation, cleaning and reformatting datasets, supporting qualitative and thematic data analysis, and interpreting results.
Additionally, we used generative AI tools for brainstorming.
We have reviewed all AI-assisted work.
All LLM assisted proofs were checked line by line by the authors.
We take responsibility for the final content of this work, including text, claims, or artifacts produced with the aid of generative AI.

\bibliographystyle{plainnat}
\bibliography{main}

\section*{Checklist}

\begin{enumerate}

    \item For all models and algorithms presented, check if you include:
    \begin{enumerate}
        \item A clear description of the mathematical setting, assumptions, algorithm, and/or model. [Yes]
        \item An analysis of the properties and complexity (time, space, sample size) of any algorithm. [Not Applicable]
        \item (Optional) Anonymized source code, with specification of all dependencies, including external libraries. [Yes]
    \end{enumerate}
    
    \item For any theoretical claim, check if you include:
    \begin{enumerate}
        \item Statements of the full set of assumptions of all theoretical results. [Yes]
        \item Complete proofs of all theoretical results. [Yes]
        \item Clear explanations of any assumptions. [Yes]
    \end{enumerate}

    \item For all figures and tables that present empirical results, check if you include:
    \begin{enumerate}
        \item The code, data, and instructions needed to reproduce the main experimental results (either in the supplemental material or as a URL). [Yes]
        \item All the training details (e.g., data splits, hyperparameters, how they were chosen). [Yes]
        \item A clear definition of the specific measure or statistics and error bars (e.g., with respect to the random seed after running experiments multiple times). [Yes]
        \item A description of the computing infrastructure used. (e.g., type of GPUs, internal cluster, or cloud provider). [Yes]
    \end{enumerate}
    
    \item If you are using existing assets (e.g., code, data, models) or curating/releasing new assets, check if you include:
    \begin{enumerate}
        \item Citations of the creator If your work uses existing assets. [Yes]
        \item The license information of the assets, if applicable. [Yes]
        \item New assets either in the supplemental material or as a URL, if applicable. [Yes]
        \item Information about consent from data providers/curators. [Not Applicable]
        \item Discussion of sensible content if applicable, e.g., personally identifiable information or offensive content. [Not Applicable]
    \end{enumerate}
    
    \item If you used crowdsourcing or conducted research with human subjects, check if you include:
    \begin{enumerate}
        \item The full text of instructions given to participants and screenshots. [Not Applicable]
        \item Descriptions of potential participant risks, with links to Institutional Review Board (IRB) approvals if applicable. [Not Applicable]
        \item The estimated hourly wage paid to participants and the total amount spent on participant compensation. [Not Applicable]
    \end{enumerate}
    
\end{enumerate}

\clearpage
\appendix

\counterwithin{table}{section}
\setcounter{table}{0}
\counterwithin{figure}{section}
\setcounter{figure}{0}
\counterwithin{theorem}{section}
\setcounter{theorem}{0}
\counterwithin{corollary}{section}
\setcounter{corollary}{0}

\onecolumn
\section*{Appendix}
\section{CLOSED-FORM ELBO}
\label{app:closed-form_elbo}

\begin{align}
    \mathbb{E}_{q(w)}\left[ \log p(y | w) \right] &= \textstyle -\frac{1}{2} \left[ N \log (2\pi\sigma_n^2) + \frac{1}{\sigma_n^2} \left\| y - \sigma_f \Phi m \right\|_2^2 + \frac{\sigma_f^2}{\sigma_n^2} \tr(\Phi^\top \Phi S) \right] \\
    -\KL{q(w)}{p(w)} &= \textstyle -\frac{1}{2} \left[ \tr(\Sigma_p^{-1} S) + (\mu_p - m)^\top \Sigma_p^{-1} (\mu_p - m) - R + \log \frac{\det(\Sigma_p)}{\det(S)} \right]
\end{align}

\section{COORDINATE ASCENT UPDATES}
\label{app:coordinate_ascent_updates}

\subsection{Full-Rank Covariance}

\begin{align}
    \frac{\partial J}{\partial m} &= - \frac{\sigma_f}{\sigma_n^2} \Phi^\top (\sigma_f \Phi m - y) - \frac{1}{\tau} m \\
    m^* &= \left( \frac{\sigma_f^2}{\sigma_n^2} \Phi^\top \Phi + \frac{1}{\tau} I_R \right)^{-1} \frac{\sigma_f}{\sigma_n^2} \Phi^\top y
\end{align}

\begin{align}
    \frac{\partial J}{\partial S} &= - \frac{1}{2} \left( \frac{\sigma_f^2}{\sigma_n^2} \Phi^\top \Phi + \frac{1}{\tau} I_R + S^{-1} \right) \\
    S^* &= \left( \frac{\sigma_f^2}{\sigma_n^2} \Phi^\top \Phi + \frac{1}{\tau} I_R \right)^{-1}
\end{align}

\begin{align}
    \frac{\partial J}{\partial \sigma_n^2} &= - \frac{1}{2} \left( \frac{N}{\sigma_n^2} - \frac{1}{\sigma_n^4} \left( \left\| y - \sigma_f \Phi m \right\|_2^2 + \sigma_f^2 \tr(\Phi^\top \Phi S) \right) \right) \\
    (\sigma_n^2)^* &= \frac{1}{N} \left( \left\| y - \sigma_f \Phi m \right\|_2^2 + \sigma_f^2 \tr(\Phi^\top \Phi S) \right)
\end{align}

\begin{align}
    \frac{\partial J}{\partial \sigma_f} &= \frac{1}{\sigma_n^2} \left( y^\top \Phi m - \sigma_f m^\top \Phi^\top \Phi m - \sigma_f \tr(\Phi^\top \Phi S) \right) \\
    \sigma_f^* &= \frac{y^\top \Phi m}{m^\top \Phi^\top \Phi m + \tr(\Phi^\top \Phi S)}
\end{align}

\begin{align}
    \frac{\partial J}{\partial \tau} &= -\frac{1}{2} \left( -\frac{1}{\tau^2} \tr(S) - \frac{1}{\tau^2} \left\| m \right\|_2^2 + \frac{R}{\tau} \right) \\
    \label{eq:tau}
    \tau^* &= \frac{ \tr(S) + \left\| m \right\|_2^2}{R}
\end{align}

\subsection{Diagonal Covariance}

\begin{align}
    \frac{\partial J}{\partial s_r} &= - \frac{1}{2} \left( \frac{\sigma_f^2}{\sigma_n^2} (\Phi^\top \Phi)_{rr} + \frac{1}{\tau} - \frac{1}{s_r} \right) \\
    \hat{s}_r &= \frac{1}{\frac{\sigma_f^2}{ \sigma_n^2} (\Phi^\top \Phi)_{rr} + \frac{1}{\tau}}
\end{align}

\subsection{Low-Rank Plus Diagonal Covariance}

% \begin{align}
%     \tr(\Phi^\top \Phi S) = \tr(\Phi_1^\top \Phi_1 L L^\top) + \tr(\Phi_1^\top \Phi_2 C L L^\top) + \tr(\Phi_2^\top \Phi_1 L L^\top C^\top) + \tr(\Phi_2^\top \Phi_2 C L L^\top C^\top) + \tr(\Phi_2^\top \Phi_2 \diag(d))
% \end{align}

% \begin{align}
%     S = \begin{bmatrix} S_{11} & S_{12} \\ S_{21} & S_{22} \end{bmatrix} = \begin{bmatrix} L L^\top & L L^\top C^\top \\  C L L^\top & C L L^\top C^\top + \diag(d) \end{bmatrix}
% \end{align}

% \begin{align}
%     \tr(S) &= \tr(S_{11}) + \tr(S_{22}) && \text{(trace of block matrix)} \\
%     &= \tr(L L^\top) + \tr(C L L^\top C^\top + \diag(d)) && \text{(definition)} \\
%     &= \tr(L L^\top) + \tr(C L L^\top C^\top) + \tr(\diag(d)) && \text{(additivity of trace)}
% \end{align}

% \begin{align}
%     \det(S) &= \det(S_{11}) \det(S_{22} - S_{21} S_{11}^{-1} S_{12}) && \text{(Schur's formula)} \\
%     &= \det(LL^\top) \det((CLL^\top C^\top + \diag(d)) - (CLL^\top)(LL^\top)^{-1}(LL^\top C^\top)) && \text{(definition)} \\
%     &= \det(LL^\top) \det(\diag(d)) && \text{(canceling)}
% \end{align}

\begin{align}
    \frac{\partial J}{\partial C} &= -\frac{\sigma_f^2}{\sigma_n^2}\Phi_2^\top(\Phi_1 + \Phi_2 C)LL^\top - \frac{1}{\tau}CLL^\top \\
    \hat{C} &= -\left(\frac{\sigma_f^2}{\sigma_n^2}\Phi_2^\top\Phi_2 + \frac{1}{\tau}I\right)^{-1}\frac{\sigma_f^2}{\sigma_n^2}\Phi_2^\top\Phi_1
\end{align}

\begin{align}
    \frac{\partial J}{\partial L} &= - M L + (L^{-1})^\top \\
    \hat{L} &= \Chol(M^{-1})
\end{align}
where $M = \frac{\sigma_f^2}{\sigma_n^2}(\Phi_1+\Phi_2 C)^\top(\Phi_1+\Phi_2 C) + \frac{1}{\tau}(I + C^\top C)$

\begin{align}
    \frac{\partial J}{\partial d_r} &= -\frac{1}{2}\left(\frac{\sigma_f^2}{\sigma_n^2}(\Phi_2^\top\Phi_2)_{rr} + \frac{1}{\tau} - \frac{1}{d_r} \right) \\
    \hat{d}_r &= \frac{1}{\frac{\sigma_f^2}{ \sigma_n^2} (\Phi_2^\top \Phi_2)_{rr} + \frac{1}{\tau}}
\end{align}

\section{PROOFS OF KEY RESULTS}
\label{app:underestimating_proof}

\subsection{
Proof of Thm.~\ref{theorem:SNRtoMSE}: Lower SNR means Higher MSE}
\label{app:proof_SNR_to_MSE}

Recall Thm.~\ref{theorem:SNRtoMSE} from the main paper:

\thmSNRtoMSE* 

\textit{Proof.} For convenience, we define $\hat{\rho} 
= \frac{\hat{\sigma}^2_f}{\hat{\sigma}^2_n} > 0$.
Similarly, let $\rho^* = \frac{(\sigma_f^*)^2}{(\sigma^*_n)^2} > 0$.
By assumption, we know that $\hat{\rho} \leq \rho^*$.

First, we substitute the the CA-VEM update for $m^*$ from~Eq.~\eqref{eq:cavi_sigma} and apply the push-through identity to express the residual errors in terms of the kernel matrix $\Phi \Phi^\top$.
\begin{align}
    \left\| y - \hat{\sigma}_f \Phi m^*(\hat{\sigma}) \right\|_2^2 &\ge \left\| y - \sigma^*_f \Phi m^*(\sigma^*) \right\|_2^2 \\
    \left\| y - \Phi \left( \hat{\rho} \Phi^\top \Phi + I_R \right)^{-1} \hat{\rho} \Phi^\top y \right\|_2^2 &\ge \left\| y - \Phi \left( \rho^* \Phi^\top \Phi + I_R \right)^{-1} \rho^* \Phi^\top y \right\|_2^2 && \text{(definition)} \\
    \left\| \left( I_N - \left( \hat{\rho} \Phi \Phi^\top + I_N \right)^{-1} \hat{\rho} \Phi \Phi^\top \right) y \right\|_2^2 &\ge \left\| \left( I_N - \left( \rho^* \Phi \Phi^\top + I_N \right)^{-1} \rho^* \Phi \Phi^\top \right) y \right\|_2^2 && \text{(push-through identity)} \\
    \left\| \left( \hat{\rho}  \Phi \Phi^\top + I_N \right)^{-1} y \right\|_2^2 &\ge \left\| \left(  \rho^*  \Phi \Phi^\top + I_N \right)^{-1} y \right\|_2^2 && \text{(simplifying)}
\end{align}
Second, we apply the eigendecomposition of the $N \times N$ kernel matrix $\Phi \Phi^\top = U \Lambda U^\top$ to obtain a sum over nonnegative eigenvalues $\lambda_i$ for $i$ from 1 to $N$.
\begin{align}
    \left\| (\hat{\rho} U \Lambda U^\top + U U^\top)^{-1} y \right\|_2^2 
    &\ge \left\| (  \rho^* U \Lambda U^\top + U U^\top)^{-1} y \right\|_2^2 \\
    \left\| U (\hat{\rho} \Lambda + I_N)^{-1} U^\top y \right\|_2^2 &\ge \left\| U ( \rho^* \Lambda + I_N)^{-1} U^\top y \right\|_2^2 \\
    \intertext{We can drop the first $U$, because multiplying a vector by an orthogonal matrix does not change its $\ell_2$ norm. Simplifying what remains, we have}
    \sum_{i=1}^N \frac{(U^\top y)_i^2}{(1 + \hat{\rho} \lambda_i)^2} &\ge \sum_{i=1}^N \frac{(U^\top y)_i^2}{(1 +  \rho^* \lambda_i)^2}
\end{align}
Finally, the numerators are equal on both sides, so the inequality follows directly from the hyperparameters inequality.
Since $\hat{\rho} \le  \rho^*$ and $\lambda_i \ge 0$, we know the denominator on the left will be less than or equal the one on the right for all $N$ terms. Thus, the overall term on the left will be greater than or equal, which completes the proof.
%, we have $\frac{(U^\top y)_i^2}{(1 + \rho_{\mathrm{diag}} \lambda_i)^2} \ge \frac{(U^\top y)_i^2}{(1 +  \rho^* \lambda_i)^2}$ for all $i$, which completes the proof.~
$~\blacksquare$

\subsubsection{Empirical Bayes}

We can extend this proof for empirical Bayes by including the optimal prior variance in $\hat{\rho} = \frac{\hat{\sigma}_f^2 \hat{\tau}}{\hat{\sigma}_n^2} > 0$ and $\rho^* = \frac{(\sigma_f^*)^2 \tau^*}{(\sigma_n^*)^2} > 0$.
By Cor.~\ref{corollary:empirical_bayes}, we know that $\hat{\rho} \le \rho^*$ still holds.

%%%%%%%%
%%%%%%%%

\subsection{Proof of  Thm.~\ref{theorem:diag_overestimates_predvar}: Diagonal VI Overestimates Predictive Variance}
\label{app:proof_diag_overestimates_predvar}

Recall Thm.~\ref{theorem:diag_overestimates_predvar} from the main paper:

\diagUnderestimatesPredvar*

% \begin{theorem}[Diagonal VI overestimates the training set noise-free predictive variance]
%     Let $s_r^*$ be the optimal diagonal approximate posterior variance from Theorem~\ref{theorem:optimal_s_for_diag}. Then the ELBO overestimates the training set noise-free predictive variance
%     \begin{align}
%         \tr(\Phi^\top \Phi \diag(s^*)) \ge \tr(\Phi^\top \Phi \Sigma^*).
%     \end{align}
% \end{theorem}
    
\textit{Proof.}
For convenience, we define $\rho = \frac{\sigma^2_f}{\sigma^2_n} > 0$ as the signal variance divided by the noise variance.
Let $d = (d_1, \dots, d_R)$ and $\lambda = (\lambda_1, \dots, \lambda_R)$ denote the diagonal entries and eigenvalues of $\Phi^\top \Phi$ arranged in nonincreasing order.

We can rewrite the left-hand side (LHS) of the inequality as a sum of diagonal entries of $\Phi^\top \Phi$.
\begin{align}
    \tr(\Phi^\top \Phi \diag(\hat{s})) &= \sum_{r=1}^R d_r \hat{s}_r && \text{(trace of product with diagonal matrix)} \\
    &= \sum_{r=1}^R \frac{d_r}{\Lambda_{rr}^*} && \text{(Thm.~\ref{theorem:optimal_s_for_diag})} \\
    &= \sum_{r=1}^R \frac{d_r}{1 + \rho d_r} && \text{(definition)}
\end{align}
We can rewrite the right-hand side (RHS) of the inequality as a sum of eigenvalues of $\Phi^\top \Phi$.
\begin{align}
    \tr(\Phi^\top \Phi \Sigma^*) &= \tr\left( \Phi^\top \Phi (I_R + \rho \Phi^\top \Phi)^{-1} \right) && \text{(definition)} \\
    &= \tr\left( U \Lambda U^\top (U (I_R + \rho \Lambda) U^\top)^{-1} \right) && \text{(eigendecomposition)} \\
    &= \tr\left( U \Lambda (I_R + \rho \Lambda)^{-1} U^\top \right) && \text{(orthogonality of eigenvectors)} \\
    &= \tr\left( \Lambda (I_R + \rho \Lambda)^{-1} \right) && \text{(cyclic property of trace)} \\
    &= \sum_{r=1}^R \frac{\lambda_r}{1 + \rho \lambda_r} && \text{(trace of diagonal matrix)}
\end{align}
First, by the Schur-Horn theorem, we know that for a Hermitian matrix like $\Phi^\top \Phi$, we have
%https://en.wikipedia.org/wiki/Schur%E2%80%93Horn_theorem
\begin{align}
    \sum_{i=1}^r d_i &\le \sum_{i=1}^r \lambda_i, \quad \text{for $r = 1, \dots, R-1$}, \qquad \text{and} \qquad \sum_{i=1}^R d_i = \sum_{i=1}^R \lambda_i.
\end{align}
Second, both the LHS and RHS are sums of the same function
\begin{align}
    f(x) = \frac{x}{1 + \rho x}
\end{align}
This function is concave on $[0, \infty)$, because for any $\rho > 0$, the second derivative $f^{\prime\prime}(x) = - \frac{2\rho}{(1 + \rho x)^3}$ is always negative for $x \ge 0$.

Finally, by Karamata's inequality~\citep{karamata1932inegalite}, a generalization of Jensen's inequality, we have
%https://en.wikipedia.org/wiki/Karamata%27s_inequality
\begin{align}
    \sum_{r=1}^R f(d_r) &\ge \sum_{r=1}^R f(\lambda_r)
\end{align}
or equivalently, $\tr(\Phi^\top \Phi \diag(\hat{s})) \ge \tr(\Phi^\top \Phi \Sigma^*)$.~$\blacksquare$

\subsection{Proof of  Thm.~\ref{theorem:optimal_param_for_LRD}: Optimal Covariance Parameters for LRD VI}
\label{app:proof_optimal_param_for_LRD}

Recall Thm.~\ref{theorem:optimal_param_for_LRD} from the main paper:

\optimalParamForLRD*

\textit{Proof.}

\begin{align}
    q\left(\begin{bmatrix} w_1 \\ w_2 \end{bmatrix}\right) = \mathcal{N} \left( \begin{bmatrix} m_1 \\ m_2 \end{bmatrix}, \begin{bmatrix} LL^\top & LL^\top C^\top \\ CLL^\top & C L L^\top C^\top + \diag(d) \end{bmatrix} \right)
\end{align}

The approximate posterior factorizes as $q(w) = q(w_1) q(w_2 | w_1)$ where
\begin{align}
    q(w_1) = \mathcal{N}(m_1, LL^\top), \qquad q(w_2 | w_1) = \mathcal{N}(m_2 + C (w_1 - m_1), \diag(d)).
\end{align}
The true posterior factorizes as $p(w | y) = p(w_1 | y) p(w_2 | w_1, y)$ where
\begin{align}
    p(w_1 | y) = \mathcal{N}(\mu_1^*, \Sigma_{11}^*), \qquad p(w_2 | w_1, y) = \mathcal{N}(\mu_2^* + \Sigma_{21}^*(\Sigma_{11}^*)^{-1} (w_1 - \mu_1^*), \Sigma_{2|1}^*).
\end{align}
Using the chain rule, the KL divergence factorizes as
\begin{align}
    \KL{q(w)}{p(w | y)} = \KL{q(w_1)}{p(w_1 | y)} + \mathbb{E}_{q(w_1)}\left[ \KL{q(w_2 | w_1)}{p(w_2 | w_1, y)} \right].
\end{align}
The marginal KL divergence is minimized by matching the true marginal mean $m_1 = \mu_1^*$ and covariance matrix $L = \Chol(\Sigma_{11}^*)$.
However, the expected conditional KL divergence depends on the marginal distribution $q(w_1)$.
The closed-form conditional KL divergence is
\begin{align}
    &\textstyle \KL{q(w_2 | w_1)}{p(w_2 | w_1, y)} \\
    &= \frac{1}{2} \left[ \tr((\Sigma_{2|1}^*)^{-1} \diag(d)) + (\mu_{2|1} - m_{2|1})^\top (\Sigma_{2|1}^*)^{-1} (\mu_{2|1} - m_{2|1}) - (R-k) + \log \frac{\det(\Sigma_{2|1}^*)}{\det(\diag(d))} \right], \notag
\end{align}
where the conditional means are
\begin{align}
    m_{2|1} &= m_2 + C (w_1 - m_1), \\
    \mu_{2|1} &= \mu_2^* + \Sigma_{21}^*(\Sigma_{11}^*)^{-1} (w_1 - \mu_1^*).
\end{align}
Only the quadratic term of the conditional KL divergence depends on $w_1$, and this term is minimized by matching the conditional mean $m_2^* = \mu_2^*$ and $C^* = \Sigma_{21}^*(\Sigma_{11}^*)^{-1}$ regardless of the marginal distribution $q(w_1)$.

The remaining terms form the KL divergence between the diagonal conditional covariance matrix $\diag(d)$ and the true conditional covariance matrix $\Sigma_{2|1}^*$. 
By the block matrix inverse identity, the true conditional precision matrix is $(\Sigma_{2|1}^*)^{-1} = \Lambda_{22}^*$.
It follows from Thm.~\ref{theorem:optimal_s_for_diag} that the $r$-th element of the optimal diagonal conditional covariance is $d_r^* = \frac{1}{(\Lambda_{22}^*)_{rr}}$.~$\blacksquare$

\subsection{Proof of  Thm.~\ref{theorem:LRD_overestimates_predvar}: LRD VI Overestimates Predictive Variance}
\label{app:proof_LRD_overestimates_predvar}

\thmLRDOverestimatesPredVar*

\textit{Proof.}
We parameterize the low-rank plus diagonal covariance matrix as:
\begin{align}
    VV^\top + D = \begin{bmatrix} L \\ C L \end{bmatrix} \begin{bmatrix} L^\top & L^\top C^\top \end{bmatrix} + \begin{bmatrix} 0 & 0 \\ 0 & \diag(d) \end{bmatrix}
\end{align}
where $L \in \mathbb{R}^{k \times k}$ is lower triangular, $C \in \mathbb{R}^{R - k \times k}$, and $d \in \mathbb{R}^{R - k}$.

Minimizing the KL divergence between the marginal distributions of the low-rank component yields the optimal closed-form solutions for these block terms: $L L^\top = \Sigma_{11}^*$ and $C = \Sigma_{21}^* (\Sigma_{11}^*)^{-1}$.

Let $\Phi = \begin{bmatrix} \Phi_1 & \Phi_2 \end{bmatrix}$ be the block partition of the design matrix corresponding to the covariance partition.
We can expand the total predictive variance of the approximation as:
\begin{align}
    \tr(\Phi^\top \Phi S) = \tr(\Phi_1^\top \Phi_1 \Sigma_{11}^*) + \tr(\Phi_1^\top \Phi_2 \Sigma_{21}^*) + \tr(\Phi_2^\top \Phi_1 \Sigma_{12}^*) + \tr(\Phi_2^\top \Phi_2 \Sigma_{21}^* (\Sigma_{11}^*)^{-1} \Sigma_{12}^*) + \tr(\Phi_2^\top \Phi_2 \diag(d)).
\end{align}
We can expand the exact total predictive variance of the true posterior as:
\begin{align}
    \tr(\Phi^\top \Phi \Sigma^*) = \tr(\Phi_1^\top \Phi_1 \Sigma_{11}^*) + \tr(\Phi_1^\top \Phi_2 \Sigma_{21}^*) + \tr(\Phi_2^\top \Phi_1 \Sigma_{12}^*) + \tr(\Phi_2^\top \Phi_2 \Sigma_{22}^*)
\end{align}

\begin{align}
    \tr(\Phi_2^\top \Phi_2 \Sigma_{21}^* (\Sigma_{11}^*)^{-1} \Sigma_{12}^*) + \tr(\Phi_2^\top \Phi_2 \diag(d)) &\ge \tr(\Phi_2^\top \Phi_2 \Sigma_{22}^*) \\
    \tr(\Phi_2^\top \Phi_2 \diag(d)) &\ge \tr(\Phi_2^\top \Phi_2 \Sigma_{22}^*) - \tr(\Phi_2^\top \Phi_2 \Sigma_{21}^* (\Sigma_{11}^*)^{-1} \Sigma_{12}^*) \\
    \tr(\Phi_2^\top \Phi_2 \diag(d)) &\ge \tr(\Phi_2^\top \Phi_2 \Sigma_{2|1}^*)
\end{align}
By Thm.~\ref{theorem:diag_overestimates_predvar}, a diagonal VI approximation overestimates the total predictive variance. $\blacksquare$

\section{FURTHER EXPERIMENTS AND RESULTS}

\subsection{Training Setup}
\label{app:expt_setup}

We initialize hyperparameters at their data generating values.
For each covariance type (diagonal, LRD, LR, FR), we run CA-VEM for $1{,}000$ iterations, updating the variational parameters $m, S$ and hyperparameters $\sigma$ in closed form (App.~\ref{app:coordinate_ascent_updates}).
Results in Fig.~\ref{fig:varying_k} fix $\ell = 1$ while varying $k$, reporting the mean of the estimated SNR ratio (with shaded 95\% interval) over 500 samples of $N = 20$ training data.
Fig.~\ref{fig:under_overestimating} visualizes the predictions of a best estimate for $m, S, \sigma$ at fixed $k$ for one $N = 20$ dataset.
For the experiment in Fig.~\ref{fig:under_overestimating}, we learn lengthscale $\ell$ with gradient descent, interleaving first-order gradient updates to $\ell$ with CA-VEM updates.

\subsection{Results as Dataset Size Increases}
\begin{figure}[htpb!]
    \centering
    \includegraphics[width=0.5\linewidth]{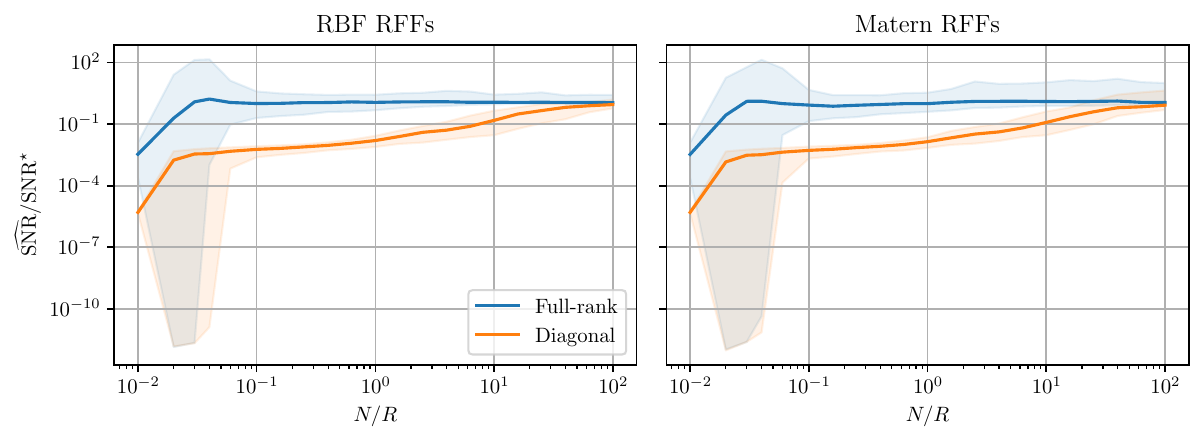}
    \caption{Relative SNR estimate vs. relative dataset size for RBF (left) and Mat\'ern (right) RFFs.
    We vary $N \in [1, 10{,}000]$ while fixing $R=100$.
    \textbf{The underestimated SNR improves as $N$ increases.}
    }%endcaption
    \label{fig:varying_N}
\end{figure}

Fig.~\ref{fig:varying_N} shows how the estimated SNR changes as dataset size $N$ increases, while holding $R$ constant. 
We see that in this regime, as $N$ increases the diagonal covariance approximate posterior gets closer and closer to estimating the SNR correctly.

\subsection{Predictive Posterior Visuals with Fixed Lengthscale}

\begin{figure*}[htpb!]
    \centering
    \includegraphics[width=\linewidth]{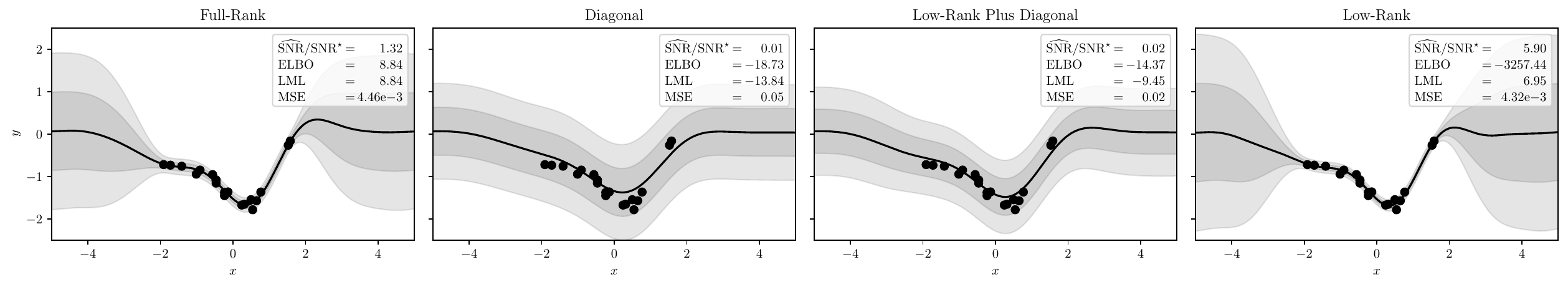}
    \caption{
    Predictive posterior $p(\ytest|y)$ for full-rank, diagonal, low-rank plus diagonal, and low-rank covariances (left to right), after estimating variational parameters and hyperparameters.
    We fix $N = 20$, $R = 1{,}024$, $k = 512$, and $\ell = 1$.
    }
    \label{fig:under_overestimating_lengthscale=1.0}
\end{figure*}
\begin{figure*}[htpb!]
    \centering
    \includegraphics[width=\linewidth]{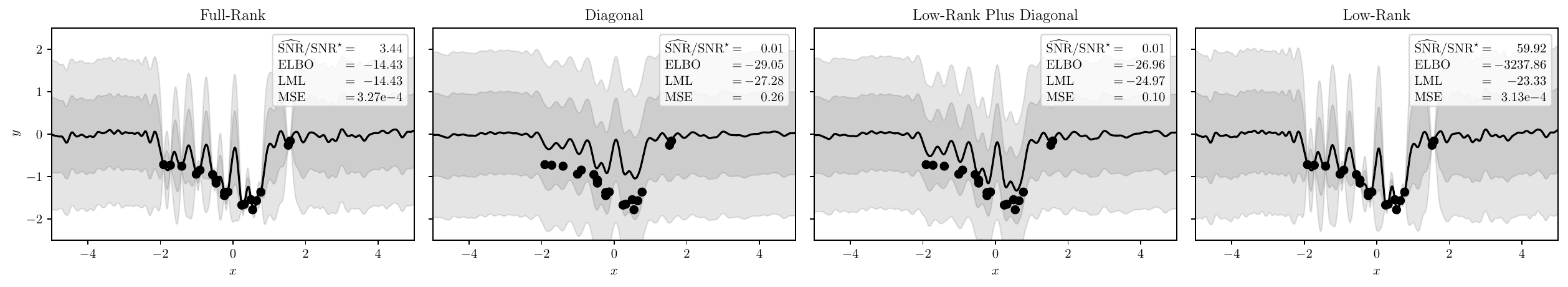}
    \caption{
    Predictive posterior $p(\ytest|y)$ for full-rank, diagonal, low-rank plus diagonal, and low-rank covariances (left to right), after estimating variational parameters and hyperparameters.
    We fix $N = 20$, $R = 1{,}024$, $k = 512$, and $\ell = 0.1$.
    }
    \label{fig:under_overestimating_lengthscale=0.1}
\end{figure*}

%\input{fig_varying_R}

%In contrast, increasing $R$ with $N$ fixed widens the gap.
% This is consistent with \citet{coker2022wide}'s finding that sufficiently wide mean-field BNNs increasingly ignore the data.

% \input{fig_varying_noise}

% Fig.~\ref{fig:varying_noise} shows how the noise variance affects the learned SNR.
% At low noise, the full-rank covariance recovers the SNR, while the diagonal covariance underestimates the SNR.
% However, at high noise both the full-rank and diagonal covariance fail drastically.
% The instability at high noise comes from a critical point in the coordinate ascent update for the signal amplitude: at high noise the update is approximately $\sigma_f^{(t)} \leftarrow \left( \frac{y^\top \Phi \Phi^\top y}{\sigma_n^2 \tr(\Phi^\top \Phi)} \right) \sigma_f^{(t-1)}$, where $\mathbb{E}_{p(y)}\left[\frac{y^\top \Phi \Phi^\top y}{\sigma_n^2 \tr(\Phi^\top \Phi)}\right]=1$, so whether a given dataset's signal amplitude collapses toward zero or grows to a nontrivial fit is close to a coin flip rather than an optimization failure.

\newpage
\subsection{Predictive Posterior Visuals Using Tempering}
\begin{figure*}[htpb!]
    \centering
    \includegraphics[width=\linewidth]{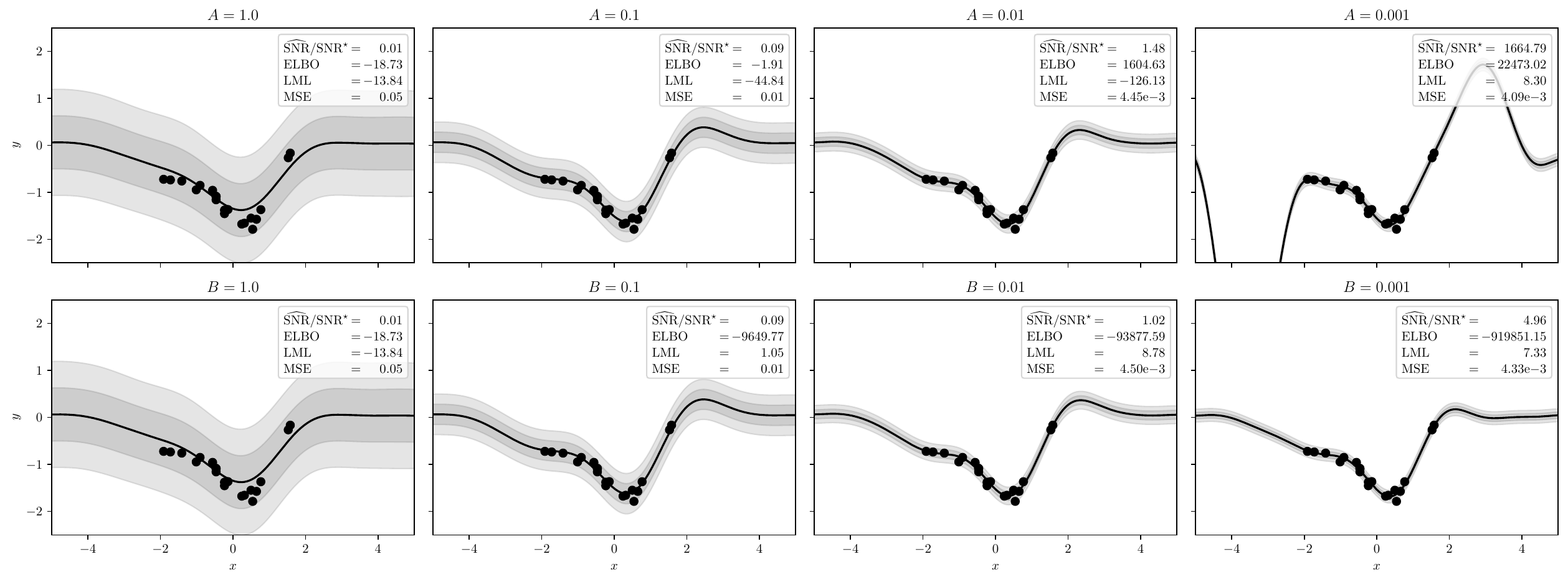}
    \caption{
    Predictive posterior $p(\ytest|y)$ for tempered likelihood (top row) and joint (bottom row), after estimating variational parameters and hyperparameters.
    We fix $N = 20$ and $R = 1{,}024$.
    }%endcaption
    \label{fig:tempered_likelihood_and_joint}
\end{figure*}

\section{EMPIRICAL BAYES}

\subsection{Proof of Thm.~\ref{theorem:diag_underestimates_postvar}: Diagonal VI Overestimates Total Posterior Variance}

\begin{theorem}[Diagonal VI underestimates the total posterior variance]
    \label{theorem:diag_underestimates_postvar}
    Let $\hat{s}_r$ be the optimal diagonal approximate posterior variances from Thm.~\ref{theorem:optimal_s_for_diag}. Then the ELBO underestimates the total posterior variance
    \begin{align}
        \tr(\diag(\hat{s})) \le \tr(\Sigma^*).
    \end{align}
\end{theorem}

\textit{Proof.}
First, we partition $\Lambda^*$ into the $r$-th component and its complement $\neg r$
\begin{align}
    \Lambda^* =
    \begin{pmatrix}
    \Lambda^*_{rr} & \Lambda^*_{r,\neg r} \\
    \Lambda^*_{\neg r,r} & \Lambda^*_{\neg r,\neg r}
    \end{pmatrix}.
\end{align}
Second, by the block matrix inverse identity we have
\begin{align}
    \Sigma_{rr}^* = \left( \Lambda_{rr}^* - \Lambda_{r,\neg r}^* (\Lambda_{\neg r,\neg r}^*)^{-1} \Lambda_{\neg r,r}^* \right)^{-1} && \text{(see Eq.~(399) in \citet{petersen2008matrix}).}
\end{align}
Since $\Lambda^*$ is positive definite (PD), the principal submatrix $\Lambda^*_{\neg r,\neg r}$ is PD, and therefore its inverse $(\Lambda^*_{\neg r,\neg r})^{-1}$ is PD.
Since the quadratic form $\Lambda^*_{r,\neg r} (\Lambda^*_{\neg r,\neg r})^{-1} \Lambda^*_{\neg r,r} \ge 0$, it follows that
\begin{align}
    \Sigma_{rr}^* \ge \frac{1}{\Lambda_{rr}^*} = \hat{s}_r.
\end{align}
Finally, by summing over $r=1, \dots, R$ we have: $\tr(\diag(\hat{s})) \le \tr(\Sigma^*)$.~$\blacksquare$

\begin{corollary}[Diagonal VI underestimates the prior variance]
    \label{corollary:empirical_bayes}
    Given hyperparameters $\sigma$ and mean $m$, perform CA-VEM updates to the covariance parameters and then $\tau$. The resulting value (denoted with hat) for diagonal VI is bounded relative to the corresponding value (star) using FR updates:
    \begin{align}
        \hat{\tau} \le \tau^*,
    \end{align}
    where equality holds only if $\Sigma^*$ is a diagonal matrix.
\end{corollary}

\textit{Proof.}
We update $\tau$ by plugging in the diagonal and full-rank covariance into the $\tau$ update.
Comparing the former to the latter, we find
\begin{align}
    \frac{\tr(\diag(\hat{s})) + \left\| m \right\|_2^2}{R} \le \frac{ \tr(\Sigma^*) + \left\| m \right\|_2^2}{R}.
\end{align}
which holds by the inequality in Thm.~\ref{theorem:diag_underestimates_postvar}, as long as $m$ is the same on both sides.

\end{document}